\documentclass[sigconf]{acmart}

\usepackage{balance}

\DeclareMathOperator*{\argmin}{arg\,min}

\usepackage{multirow}

\def\BibTeX{{\rm B\kern-.05em{\sc i\kern-.025em b}\kern-.08emT\kern-.1667em\lower.7ex\hbox{E}\kern-.125emX}}
    
\copyrightyear{2019}
\acmYear{2019}
\setcopyright{acmcopyright}
\acmConference[SIGIR '19] {42nd International ACM SIGIR Conference on Research and Development in Information Retrieval}{July 21--25, 2019}{Paris, France}
\acmBooktitle{42nd Int'l ACM SIGIR Conference on Research \& Development in Information Retrieval (SIGIR'19), July 21--25, 2019, Paris, France}
\acmPrice{15.00}
\acmDOI{10.1145/XXXXXX.XXXXXX}
\acmISBN{978-1-4503-6172-9/19/07}

\begin{document}

\title{Interact and Decide: Medley of Sub-Attention Networks for Effective Group Recommendation}


\author{Lucas Vinh Tran}
\affiliation{
  \institution{Nanyang Technological University}
  \institution{Institute for Infocomm Research, A*STAR}
}
\email{trandang001@e.ntu.edu.sg}

\author{Tuan-Anh Nguyen Pham}
\affiliation{
  \institution{Nanyang Technological University}
}
\email{pham0070@e.ntu.edu.sg}

\author{Yi Tay}
\affiliation{
  \institution{Nanyang Technological University}
}
\email{ytay017@e.ntu.edu.sg}
 
\author{Yiding Liu}
\affiliation{
  \institution{Nanyang Technological University}
}
\email{liuy0130@e.ntu.edu.sg}

\author{Gao Cong}
\affiliation{
  \institution{Nanyang Technological University}
}
\email{gaocong@e.ntu.edu.sg}

\author{Xiaoli Li}
\affiliation{
  \institution{Institute for Infocomm Research, A*STAR}
}
\email{xlli@i2r.a-star.edu.sg}

%

%
\begin{abstract}
This paper proposes \textit{Medley of Sub-Attention Networks} (MoSAN), a new novel neural architecture for the \textit{group} recommendation task. Group-level recommendation is known to be a challenging task, in which intricate group dynamics have to be considered. As such, this is to be contrasted with the standard recommendation problem where recommendations are personalized with respect to a single user. Our proposed approach hinges upon the key intuition that the decision making process (in groups) is generally dynamic, i.e., a user's decision is highly dependent on the other group members. All in all, our key motivation manifests in a form of an attentive neural model that captures fine-grained interactions between group members. In our MoSAN model, each sub-attention module is representative of a single  member, which models a user's preference with respect to all other group members. Subsequently, a Medley of Sub-Attention modules is then used to collectively make the group's final decision. Overall, our proposed model is both expressive and effective. Via a series of extensive experiments, we show that MoSAN not only achieves state-of-the-art performance but also improves standard baselines by a considerable margin.
\end{abstract}

%
%


%

\copyrightyear{2019} 
\acmYear{2019} 
\setcopyright{acmcopyright}
\acmConference[SIGIR '19]{Proceedings of the 42nd International ACM SIGIR Conference on Research and Development in Information Retrieval}{July 21--25, 2019}{Paris, France}
\acmBooktitle{Proceedings of the 42nd International ACM SIGIR Conference on Research and Development in Information Retrieval (SIGIR '19), July 21--25, 2019, Paris, France}
\acmPrice{15.00}
\acmDOI{10.1145/3331184.3331251}
\acmISBN{978-1-4503-6172-9/19/07}

%

%
\maketitle

\section{Introduction}

People often participate in activities in groups, e.g., having dinners with colleagues, watching movies with partners, and shopping with friends. This calls for effective techniques for group recommendation. Unfortunately, existing recommendation algorithms designed for individuals are not effective for group recommendation. The availability of group event data further promotes the research interest on how to make effective recommendations for a group of users \cite{DBLP:journals/pvldb/Amer-YahiaRCDY09,DBLP:conf/www/CarvalhoM13a,DBLP:conf/www/GorlaLR013,DBLP:conf/cikm/LiuTYL12,DBLP:conf/sigir/YeLL12,DBLP:conf/kdd/YuanCL14,Koren:2009:MFT:1608565.1608614,DBLP:conf/recsys/BaltrunasMR10,DBLP:conf/ecscw/OConnorCKR01,DBLP:conf/flairs/McCarthySCMSN06} which facilitates groups making decisions, and helps social network services improve user engagement. This paper is concerned with designing highly effective recommender systems that are targeted at modeling group preferences as opposed to individual preferences, where groups are ad-hoc (any combination of individuals) rather than pre-defined.


Group preferences are not straightforward to model, given the inherent complexity of group dynamics. To this end, this work aims to exploit the interactions between group members in order to drive the model towards highly effective group-level recommendations. Moreover, it is natural that collective decisions have a tendency to be dynamic, i.e., a user's preference may be highly influenced by the other members in the group. Group-level agreement tends to require a \textit{consensus} amongst group members, in which this consensus largely depends on each member's roles and expertise.\footnote{While this is not explicitly captured with any semantic information or meta-data, we hypothesize that this can be implicitly captured with simply interaction data.} As such, it is crucial to model the \textit{interactions} among group members. However, existing proposals for group recommendation fail to model the interactions of group members well. Most of existing solutions belong to memory based methods that are based on either preference aggregation \cite{Koren:2009:MFT:1608565.1608614} or score aggregation strategy \cite{DBLP:conf/recsys/BaltrunasMR10,DBLP:conf/ecscw/OConnorCKR01} and do not consider the interactions of group members. These strategies overlook the interactions between group members, and use trivial methods to aggregate members' preferences. Some existing solutions are model-based approaches and try to exploit user interactions for group recommendation. However, they cannot fully utilize the user interactions as to be discussed in Section \ref{sec:related_work}.


In our work, to model the interactions among group members, we propose a new neural architecture for group recommendation. Specifically, our architecture is a new variant of the attention mechanism in which each group member is represented with a single sub-attention network. Subsequently, a group of users is then represented as a \textit{`medley'} of sub-attention networks that is responsible for making the overall recommendation. The role of each sub-attention network is to capture the preference of its representative group member, \textit{with respect to} all other members in the group. As such, our proposed model leverages user-user interactions for making group recommendation decisions, and is not only well-aligned with the fundamental intuition of group-level dynamics but also expressive in the sense that it considers the user-user interactions. In fact, our experiments demonstrate that a simple attentive aggregation of user representations is insufficient and has roughly identical performance to that of an average pooling matrix factorization (MF) baseline (More details will be discussed in Section \ref{sec:experiments}). On the other hand, our experiments show that our model is significantly better than seven state-of-the-art baselines. All in all, our core intuition serves as an inductive bias for our model, providing more effective group recommender performance on multiple benchmark datasets. 

\smallskip\noindent \textbf{Our Contributions. }
Overall, the key contributions of this work are summarized as follows:

\begin{itemize}
  \item We propose MoSAN  (\textit{Medley of Sub-Attention Networks}), a novel deep learning architecture for the group recommendation problem. Our model distinguishes itself from all prior work in group recommendation based on the fact that it considers user-user interactions using sub-attention networks. To the best of our knowledge, this is the first neural model that explores the usage of user-user interactions for the group recommendation task.
  \item We conduct extensive experiments on four publicly available benchmark datasets. Our experimental results demonstrate that MoSAN achieves state-of-the-art performance, outperforming a myriad of strong competitors in the task at hand.
  \item In addition to comparison against well-studied baselines, we conduct ablation studies against two baselines AVG-MF (Average Matrix Factorization) and ATT-AVG (Attentive Aggregation) and observe that our approach significantly outperforms both approaches. This shows that our proposed model provides a more useful inductive bias for the task at hand. 
  \item We show that the attention weights of MoSAN are interpretable, i.e., it is able to discover the different weights of each user across groups, highlighting the impact of each user in different groups.
\end{itemize}

\section{Related Work}
\label{sec:related_work}

\noindent \textbf{Group Recommendation Systems. } Group recommendation methods can be characteristically dichotomized into memory-based and model-based approaches, where the memory-based approach can be further divided into the preference aggregation and the score aggregation \cite{DBLP:journals/pvldb/Amer-YahiaRCDY09}. The preference aggregation makes recommendations based on a group profile that combines all user preferences \cite{DBLP:journals/umuai/YuZHG06}, while the score aggregation computes a score of an item for each user, and then aggregates the scores across users to derive a group recommendation score of the item \cite{DBLP:conf/recsys/BaltrunasMR10,DBLP:conf/ecscw/OConnorCKR01}. The two most popular strategies for score aggregation are the average (AVG) and the least misery (LM) strategies. The AVG strategy takes the average score across individuals in the group as the final recommendation score, thereby maximizing overall group satisfaction \cite{DBLP:journals/umuai/YuZHG06}. Alternatively, the LM strategy pleases everyone by choosing the lowest among all individuals' scores as the final score \cite{DBLP:conf/recsys/BaltrunasMR10}. Both score aggregation methods have major drawbacks. The AVG strategy may return items that are favorable to some members but not to the others, while the LM strategy may end up recommending mediocre items that no one either loves or hates. \cite{DBLP:conf/recsys/BaltrunasMR10} pointed out that the performance of either strategy depends on group size and inner-group similarity. \cite{DBLP:journals/pvldb/Amer-YahiaRCDY09} proposed the concepts of relevance and disagreement. Arguing that preference disagreements on each item among group members are inevitable, the authors experimentally show that taking into account disagreement significantly improves the recommendation quality of AVG and LM strategies. 

Model-based approaches  \cite{Koren:2009:MFT:1608565.1608614,DBLP:conf/kdd/WangB11,DBLP:journals/advai/SuK09,DBLP:conf/wsdm/AgarwalC10} for group recommendation are also notable. \cite{DBLP:conf/recsys/SekoYMM11} proposed a model that incorporates item categories into recommendation, arguing that item categories influence the group's decision and items of different categories are not strictly comparable. The method, however, only applies to pre-defined groups such as couples, which can be treated as pseudo-users and apply single user recommendation techniques, while real-life groups are often ad-hoc and formed just for one-off or few activities \cite{DBLP:conf/cikm/LiuTYL12,Quintarelli:2016:RNI:2959100.2959137}. Applying game theory in group recommendation, \cite{DBLP:conf/www/CarvalhoM13a} considered each group event as a non-cooperative game, or a game with competition among members in the group, and suggested that the recommendation goal should be the game's Nash equilibrium. However, since a Nash equilibrium can be a set of items, the game theory approach may fail to recommend one specific item.

Probabilistic models have also been applied to solve group recommendation. \cite{DBLP:conf/cikm/LiuTYL12} proposed a personal impact topic (PIT) model for group recommendation, assuming that the most influential user should represent the group and have big impact on the group's decisions. However, such an assumption does not reflect the reality that a user's influence only contributes to the group's final decision if he/she is an expert in the field. \cite{DBLP:conf/kdd/YuanCL14} proposed a consensus model (COM) for group recommendation. The model assumes (i) that a user's influence depends on the topic of decision, and (ii) that the group decision making process is subject to both the topic of the group's preferences and each user's personal preferences. Despite such assumptions, COM suffers from a drawback similar to that of PIT: COM assumes that a user has the same probability to follow the group's decisions across different groups. Additionally, \cite{DBLP:conf/www/GorlaLR013} assumed that the score of a candidate item depends not only on its relevance to each member in a group but also its relevance to the whole group. They develop an information-matching based model for group recommendation, but the model suffers from high time complexity, taking several days to run on several datasets as reported by \cite{DBLP:conf/kdd/YuanCL14}. Owing to its computational prohibitivity, we do not compare with the method in our experiments.

Recently, \cite{DBLP:conf/aaai/HuCXCGC14} proposed a deep-architecture model called DLGR that learns high-level comprehensive features of group preferences to avoid the vulnerability of the data. Similar to the work \cite{DBLP:conf/recsys/SekoYMM11}, DLGR only focuses on pre-defined groups instead of ad-hoc groups, and thus cannot be applied to our setting. Therefore, we do not compare DLGR with our proposed model in this paper.

\smallskip\noindent \textbf{Neural Recommender Systems. } Neural networks have been extensively applied in recommender systems thanks to their high-quality recommendations \cite{DBLP:conf/recsys/Cheng0HSCAACCIA16,DBLP:conf/recsys/CovingtonAS16,DBLP:conf/kdd/OkuraTOT17,DBLP:conf/www/HeLZNHC17,DBLP:conf/sigir/ChenZ0NLC17,DBLP:conf/kdd/TayLH18}. Particularly, deep learning is able to capture non-linear and non-trivial relationships between users and items, which provides in-depth understanding of user demands and item characteristics, as well as the interactions between them. A tremendous part of literature has focused on integrating deep learning into recommender systems to perform various recommendation tasks, where a comprehensive review can be found at \cite{DBLP:journals/csur/ZhangYST19}. However, there is very little work that exploits neural techniques into group recommendation.

\smallskip\noindent \textbf{Neural Attention. } Attention mechanism is one of the most exciting recent advancements in deep learning \cite{DBLP:journals/corr/abs-1810-04805, DBLP:conf/nips/VaswaniSPUJGKP17, DBLP:conf/cvpr/VinyalsTBE15,DBLP:journals/corr/BahdanauCB14,DBLP:conf/nips/ChorowskiBSCB15}. The usage of neural attention in recommender systems has also gained considerable interest, with many works that exploit this recent advance for standard recommendation tasks \cite{DBLP:conf/kdd/TayLH18,Xiao:2017:AFM:3172077.3172324,DBLP:conf/cikm/ChinZJC18}. Notably, a recent work exploits neural attention for a group recommendation setup called AGREE \cite{Cao:2018:AGR:3209978.3209998}. However, AGREE is different from our model as AGREE also focuses on pre-defined groups where it requires additional group preference representation information as one component to learn the final representation of a group. In addition, if we separate the group preference representation from AGREE, the architecture becomes completely different from the original design and it similarly becomes one of our baselines called Attentive Aggregation (ATT-AVG) model, which will be discussed in Section \ref{sec:experiments}.
Hence, their framework cannot be applied to our setting. Moreover, our overall intuition and framework (usage of sub-attention and user-user interactions) significantly distinguish our work from theirs.


\section{Our Proposed Framework}
\label{sec:attgroup}
Generally speaking, our proposed MoSAN model consists of two levels: 1) user interaction learning through sub-attention network modules in which each group member is represented with a single sub-attention network module to simulate the decision making process of other members under the influential of that group member; and 2) the medley of sub-group decisions which concludes the final decision to recommend items for the group. Next, we first present the input encoding of the group recommendation problem in Section \ref{sec:input_encoding}. We then introduce the two key components of our proposed model in Section \ref{sec:medley_of_sub_attention_networks}. Lastly, we discuss the optimization method in Section \ref{sec:optimization_and_learning}.

\subsection{Input Encoding}
\label{sec:input_encoding}
Let $\textit{U} = \{u_1, u_2,\ldots , u_M\}$ and $\textit{I} = \{i_1, i_2,\ldots , i_N\}$ be the sets of $M$ users and $N$ items, respectively. We denote a history log (i.e., training instances) as $\textit{H} = \{\langle \textit{g}_1, s_1 \rangle, \langle \textit{g}_2, s_2 \rangle,\ldots , \langle \textit{g}_L, s_L \rangle \}$, where $\textit{g}_l \subset \textit{U}$ denotes an ad-hoc group and $s_l = i_k$ denotes the selected item by the group.

Given a target group $\textit{g}_{t}$, we aim to generate a recommendation list of items that group members $\textit{u}_{\textit{g}_t,i}$ in the group $\textit{g}_{t}$ may be interested in. Note that the target group can be an ad-hoc group. The group recommendation problem can be defined as follows:

\textbf{Input:} Users \textit{U}, items \textit{I}, historical log \textit{H}, and a target group $\textit{g}_t$.

\textbf{Output:} A function that maps an item to a real value to represent the item score for the target group $f_{\textit{g}_t}: \textit{I} \rightarrow \mathbb{R}$.

For each training instance, our model accepts a list of group members (users) and an item. Each user and item are represented as one-hot vectors, which map onto a dense low-dimensional vector by looking up an user and item embedding matrix, respectively. These user and item embeddings are referred to as \textit{user-latent} and \textit{item-latent} vectors (denoted as $\textbf{u}_m$ and $\textbf{v}_j$, respectively) in the rest of the paper. Moreover, we introduce an additional user embedding matrix, referred to as \textit{user-context} embedding (denoted as $\textbf{c}_l$) which is specifically designed to denote the owner of each sub-attention network. Embedding matrices are trainable parameters of the model's architecture network and are trained end-to-end with the rest of the parameters. Finally, our model trains by pairwise ranking, which essentially requires negative item samples. We define the negative items as the items that are not selected by all members in the group.

\begin{figure}[t]
	\centering
	\includegraphics[width=0.32\textwidth]{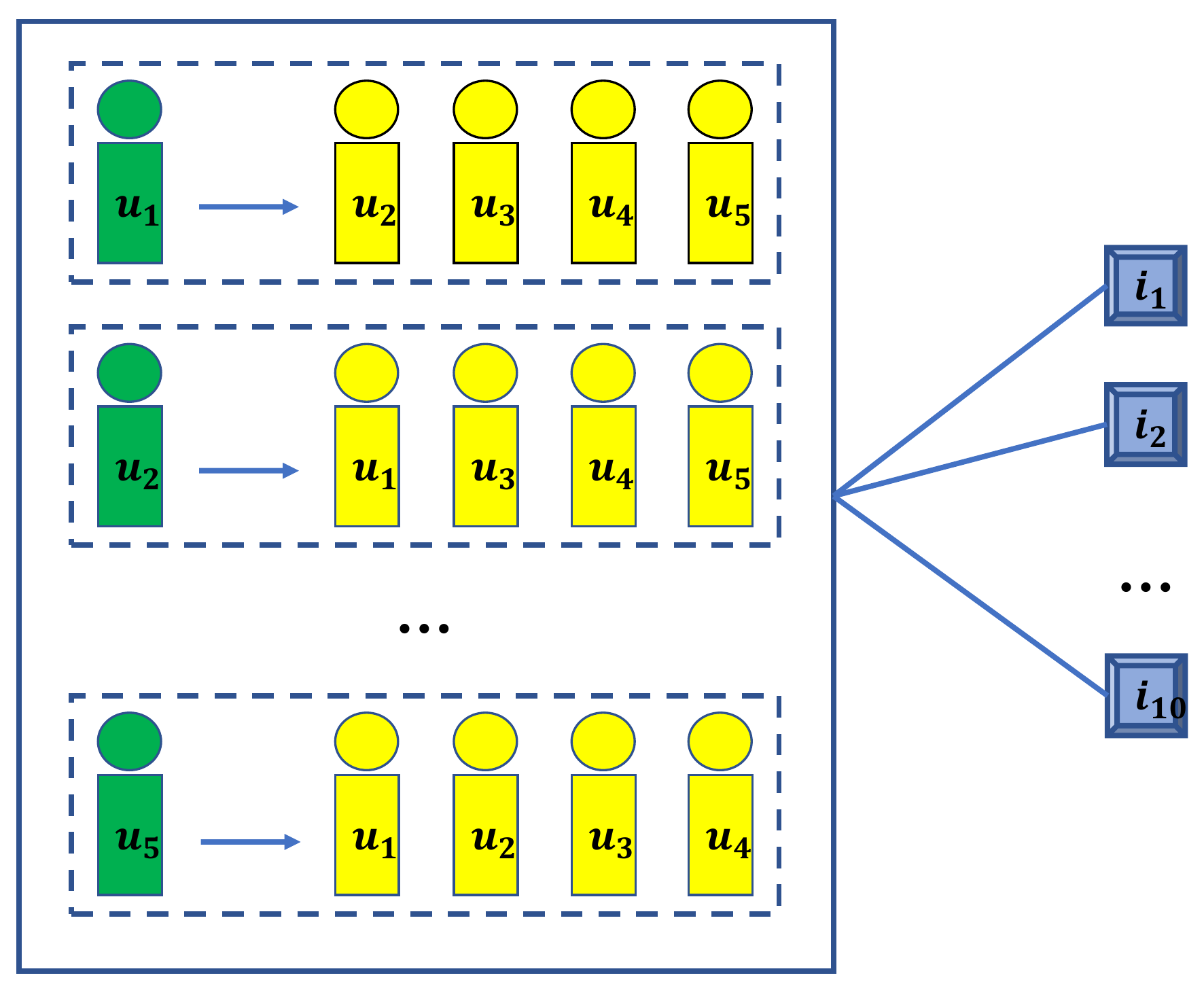}
	\caption{Illustration of group decision making process of MoSAN, in which green users represent \textit{user-context} and yellow users represent \textit{user-latent}.}
	\label{fig:input_illustration}
	\vspace{-2ex}
\end{figure}

\begin{figure*}[t]
	\centering
	\includegraphics[width=0.9\textwidth]{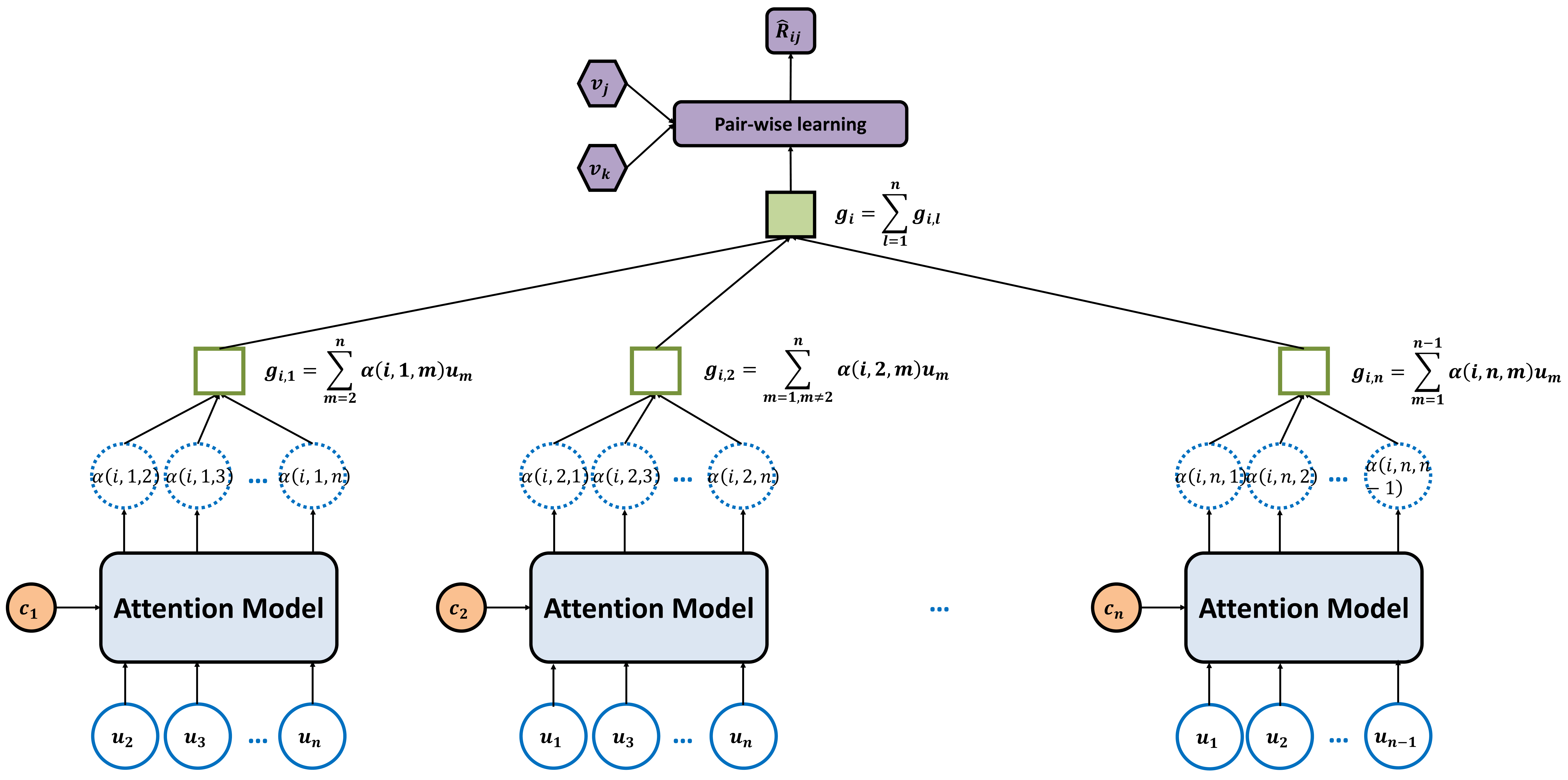}
	\caption{High level overview of our proposed Medley of Sub-Attention Networks (MoSAN) model. Each sub-attention networks is representative of a single group member, interacting with all other group members in order to learn its preference score.}
	\label{fig:model_attgroup}
\end{figure*}

Figure \ref{fig:input_illustration} illustrates the group decision making process we simulate in this paper. The solid rectangle represents the final decision of the group with each item. The inner dashed rectangles represent user interactions within the group, in which each dashed rectangle illustrates how a user (\textit{user-context}) influences other users (\textit{user-latent}) in the decision making process. 

\subsection{Medley of Sub-Attention Networks}
\label{sec:medley_of_sub_attention_networks}
This subsection introduces our proposed neural framework for group recommendation. To recapitulate, the key motivation behind our neural architecture is to enable group-level recommendations by modeling interactions between group members. Figure \ref{fig:model_attgroup} illustrates the architecture of MoSAN.

\smallskip\noindent \textbf{Motivation. } It is not straightforward to model the group preferences, given the inherent complexity of group dynamics. Therefore, typical aggregation algorithms may be insufficient for the task. Different aggregation strategies have been proposed such as average \cite{DBLP:conf/recsys/BaltrunasMR10, DBLP:conf/recsys/BerkovskyF10}, least misery \cite{DBLP:journals/pvldb/Amer-YahiaRCDY09}, or maximum satisfaction \cite{DBLP:series/sci/BorattoC11}. In general, these aggregation strategies are also known as pre-defined strategies, where they first predict the scores across individuals for candidate items, and then aggregate those predicted scores of each member in a group via the strategies to obtain the group's preferences.

We argue that these existing aggregation strategies are not sufficient to model the complexity and dynamics of the group due to its inflexibility in adjusting the weights of members in the group. For example, user $A$ may have higher impact weight than user $B$ in a given group when the group makes decision on which movie to watch, but have lower weight than user $B$ when the group makes decision on which restaurant to dine at. Recently, neural attention mechanism has been proposed as one of the most exciting advancements in deep learning \cite{DBLP:conf/cvpr/VinyalsTBE15,DBLP:journals/corr/BahdanauCB14,DBLP:conf/nips/ChorowskiBSCB15}. The concept of attention is that when people visually access an object, we tend to focus on (pay attention to) certain important parts of the object instead of the whole object in order to come up with a response. Our model adopts the attention mechanism to learn attentive and dynamic weight of each user, in which higher weights indicate the corresponding users are more important; thus, their contributions are more important for the group's final decision.

\smallskip\noindent \textbf{Our Method. } We focus on designing novel and effective neural architectures for group recommendation under representation learning framework. Specifically, we propose novel representation learning technique for learning-to-rank group-of-users and item pairs. Under the representation learning paradigm, we are able to model group representation via an end-to-end representation learning.

Let $\textbf{u}_m$ and $\textbf{v}_j$ be the embedding vector for user $m$ and item $j$, respectively. We aim to obtain an embedding vector $\textbf{g}_i$ for each group to estimate the group's preference on the item $j$. Formally, it can be defined as:

\begin{equation}
	\textbf{g}_i = f_a(\{\textbf{u}_m\}_{m \in I_i}, \textbf{v}_j)
\end{equation}
in which $\textbf{g}_i$ denotes the representation learning of group $i$ which represents its preference on item $j$; $I_i$ contains the user indexes of group $i$; and $f_a$ is the aggregation function to be specified. In MoSAN, our model architecture presenting group embedding consists of two levels: 1) Sub-Attention Network Module, and 2) the Medley of Sub-Attention Networks.

We next elaborate the two-level structure of MoSAN.

\textbf{Sub-Attention Network Module.}
For a given group $\textbf{g}_i$, we create $n$ attention sub-networks. Each attention sub-network $l$ takes the user-context\footnote{Note that the user-context vector is mainly used to differentiate the ownership of the current attention sub-network.} vector $\textbf{c}_l$ (filled orange circle) and the set of member user-latent vectors $\{\textbf{u}_1, \textbf{u}_2, \dots, \textbf{u}_{l-1}, \textbf{u}_{l+1}, \dots, \textbf{u}_n\}$ (solid blue circles) as input, and then returns the attention weight $\alpha(i, l, m)$ of each user $m$ ($m \neq l$) (dashed blue circles) of sub-network $l$ in group $\textbf{g}_i$ .

Intuitively, each attention sub-network models the interactions between each member $l$ and the rest of the group to learn the preference votes of user $l$ for other members in the group. Recall that this satisfies our key intuition and desiderata set out in the exposition of this paper. Given a user-context vector $\textbf{c}_l$ and a set of user-latent vectors $\{\textbf{u}_1, \textbf{u}_2,\ldots, \textbf{u}_n\} \setminus \{\textbf{u}_l\}$, we use a two-layer network to compute the attention score $a(i, l, m)$ as:

\begin{equation}
	a(i, l, m) = \textbf{w}^T \phi (\textbf{W}_c \textbf{c}_l + \textbf{W}_u \textbf{u}_{m, m \neq l} + \textbf{b}) + d, \quad l, m = \overline{1,n}
\end{equation}
in which the matrices $\textbf{W}_c$ and $\textbf{W}_u$ are weight matrices of the attention network
that convert user-context embedding and user-latent embedding to hidden layer respectively, and $\textbf{b}$ is the bias vector of the hidden layer; the weight vector $\textbf{w}$ and bias $d$ are the parameters of the second layer that we use to project the hidden layer to the score $a(i, l, m)$. We simply use a linear $\phi(x) = x$ as an activation function, but one can also use other functions like a ReLU function $\phi(x) = \max(0, x)$.

We normalize $a(i, l, m)$ using the Softmax function to obtain the final attention weights:

\begin{equation}
\alpha(i, l, m) = \frac{\exp(a(i, l, m))}{\sum_{m=1, m \neq l}^n \exp(a(i, l, m))}
\end{equation}
Finally, the output of each attention sub-network $l$ is calculated as the weighted sum $\textbf{g}_{i, l} = \sum \alpha(i, l, m) \textbf{u}_m$ (solid green square), which represents the group given the user-context $\textbf{c}_l$. As such, $\textbf{g}_{i, l}$ can be treated as the $l$-th representation of the group $i$, which captures the decisions of each member given that these group members' decisions are influenced by the decision of member $l$.

\textbf{Medley of Sub-Attention Networks.}
The final representation of the group $\textbf{g}_i$ is then computed as the summation $\textbf{g}_i = \sum_l \textbf{g}_{i, l}$ (filled green square). This final representation can be interpreted as the feature representation of the group of users, which can be easily matched with the item embedding to determine the recommendation score. More concretely, after we obtain the representation of the group $i$, the predicted score $\hat{R}_{ij}$ for group $i$ and item $j$ is computed as follows:
\vspace{-0.25ex}
\begin{align}
	\hat{R}_{ij} &= \textbf{g}_i^T\textbf{v}_j \nonumber \\
	&= \bigg(\sum_l \textbf{g}_{i, l}\bigg)^T\textbf{v}_j \nonumber \\
	&= \bigg(\sum_{l} \sum_{m \neq l} \alpha(i, l, m) \textbf{u}_m \bigg)^T\textbf{v}_j, \label{pred_score}
\end{align} 
in which $\textbf{v}_j$ is the item latent vector for item $j$. 

One may argue that such a simple summation could fail to consider the different number of group members, e.g., the summation of $\textbf{g}_i$ over a 3-member group tends to be much smaller than a 10-member group. As a result, according to Eq. (\ref{pred_score}), the group preference score on item $v$ of the 3-member group will be much less than that of the 10-member group due to the huge number of members, even if the 10-member group may not like this item $v$ as much as the 3-member group. However, since a large group (i.e., 10-member group) has large representation, it produces larger loss value in terms of absolute value. Hence, its gradient tend to be larger, and their group members representations are updated more during the training. If we choose appropriate learning rate, the performance will not much be affected. Moreover, in the testing phrase, we only consider and compare items to a group, so we do not have to concern about the large/small group problem. Usually the number of members in a group is small (i.e., less than 20 members), so this concern can be negligible. The normalization can also be taken into account at this layer; however, it will also be considered as an average pooling of the sub-attention networks, which makes the distinguish observation between each sub-attention network becomes less important. 

Alternatively, one could consider an additional attentive aggregation at the second layer. However, our preliminary experiments showed this yielded no improvements in performance. This is intuitive, as our model already \textit{`reasons'} over group members using the sub-attention network modules at earlier layers. Specifically, regarding the summation, the difference of each member is already captured by the sub-attention network. This is because each sub-attention network already captures the relationship of one member against the group, which aligns well with this intuition. As such, an additional attention layer did not provide any benefit to the overall network structure.

\subsection{Optimization and Learning}
\label{sec:optimization_and_learning}
\smallskip\noindent \textbf{Objective function.} Our proposed MoSAN leverages BPR \cite{DBLP:conf/uai/RendleFGS09} pair-wise learning objective to optimize the pair-wise ranking between the positive and negative items. The objective function can be rewritten as follows:

\begin{multline}
	\argmin_\Theta \sum_{(i,j,k) \in \mathcal{D}_s} - \ln \sigma\Bigg\{\bigg(\sum_{l} \sum_{m \neq l} \alpha(i, l, m) \textbf{u}_m \bigg)^T\textbf{v}_j - \\
	\bigg(\sum_{l} \sum_{m \neq l} \alpha(i, l, m) \textbf{u}_m\bigg)^T\textbf{v}_k\Bigg\} + \lambda(\|\Theta\|^2),
\end{multline}
in which $\Theta$ represents the model parameters; and $\alpha(i, l, m)$ is the weight of user $l$ votes for user $m$ in group $i$. Subsequently, the model can be trained end-to-end with an optimizer such as Adaptive Moment Estimation (Adam).

\smallskip\noindent \textbf{Learning details.} We next describe some details for learning our proposed model which are useful to replicate. 

\textbf{Mini-batch training.} We perform mini-batch training. Each mini-batch contains interactions of group members and the item adopted by the group. Specifically, we shuffle all the observed interactions, and then sample a mini-batch of those observed ones. Lastly, we form the training instances by sampling a fixed number of negative instances for each observed interaction. 

\textbf{Dropout.} We also employ dropout \cite{DBLP:journals/jmlr/SrivastavaHKSS14} to improve our proposed model's performance. Specifically, we drop with the dropout rate of $\rho$ on the first layer of our neural sub-attention network. We empirically found that applying dropout on the hidden layer of the neural attention network did boost our generalization performance. On a side note, we only apply dropout during the training phrase and disable it during the testing phase. 

\section{Experiments}
\label{sec:experiments}

In this section, we report experimental results of comparing MoSAN and seven state-of-the-art baseline techniques on four datasets. We also report the learned attention weights to evaluate the dominant users in group decision making. Such results will offer explanation for group recommendation result, which is an additional advantage of MoSAN. In general, our experiments aim to answer the following research questions (RQ):

\begin{itemize}

	\item \textbf{RQ 1:} How does MoSAN perform as compared to existing state-of-the-art methods?
	\item \textbf{RQ 2:} Are the dynamic weights learned by MoSAN more preferable than the fixed weights learned by existing methods? How effective is our attention model? What is the advantage of our attention model over the vanilla attentive aggregation baseline?
	\item \textbf{RQ 3:} How does MoSAN perform with different group sizes?

\end{itemize}

\subsection{Experimental Settings}
\label{sssec:experimental_settings}

\smallskip\noindent \textbf{Datasets.} We conduct extensive experiments on four real-world datasets. The first dataset is from an event-based social network (EBSN), Plancast,\footnote{https://www.plancast.com} which is used in \cite{DBLP:conf/kdd/LiuHTLMH12}. Plancast allows users to directly follow the event calendars of other users. An event in Plancast consists of a user group and a venue. We therefore consider an event a group, and each user in the event a group member. Members in the group will select a venue (the candidate item) to host the event. Our goal is to recommend a venue for the group event.

\begin{figure*}[t]
\begin{center}
\begin{minipage}[t]{4.39cm}
\includegraphics[width=4.39cm]{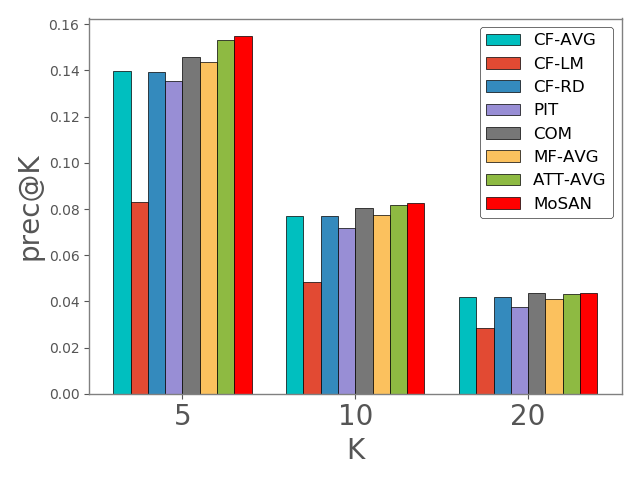}
\centering{Meetup}
\end{minipage}
\begin{minipage}[t]{4.39cm}
\includegraphics[width=4.39cm]{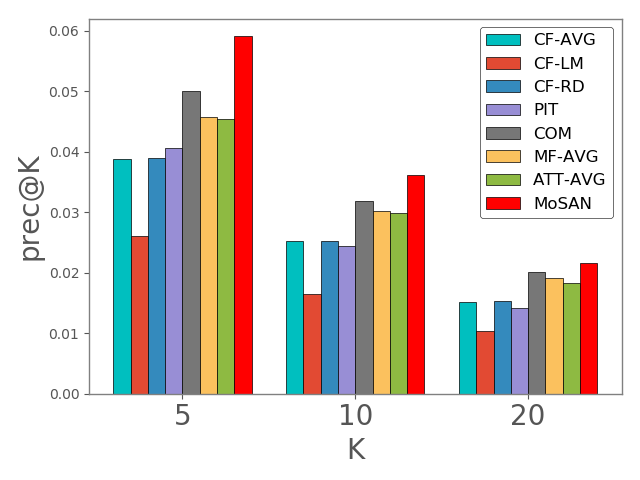}
\centering{Plancast}
\end{minipage}
\begin{minipage}[t]{4.39cm}
\includegraphics[width=4.39cm]{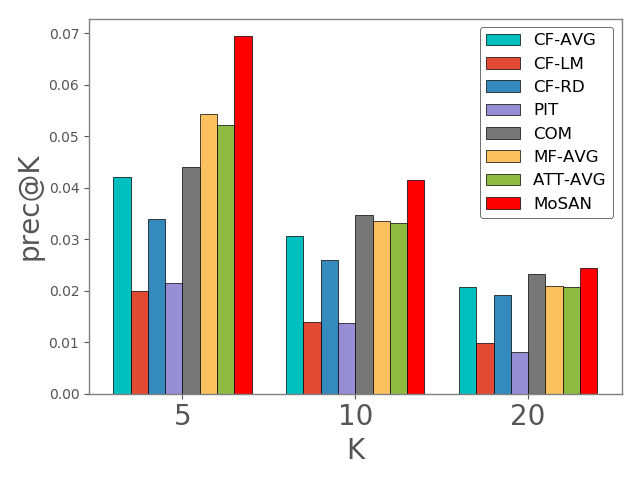}
\centering{MovieLens-Simi}
\end{minipage}
\begin{minipage}[t]{4.39cm}
\includegraphics[width=4.39cm]{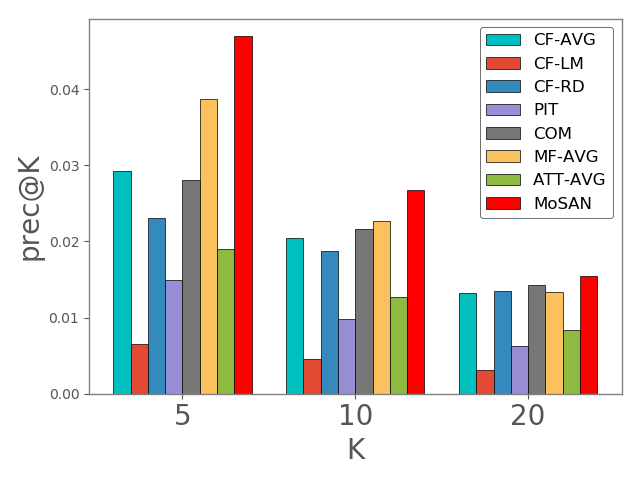}
\centering{MovieLens-Rand}
\end{minipage}
\caption{Performance of Group Recommendation Methods in terms of prec@K $(p<0.0001)$ (\textmd{\textit{Best viewed in color}}).}
\label{fig:prec}
\end{center}
\vspace{-2ex}
\end{figure*}

\begin{figure*}[t]
\begin{center}
\begin{minipage}[t]{4.39cm}
\includegraphics[width=4.39cm]{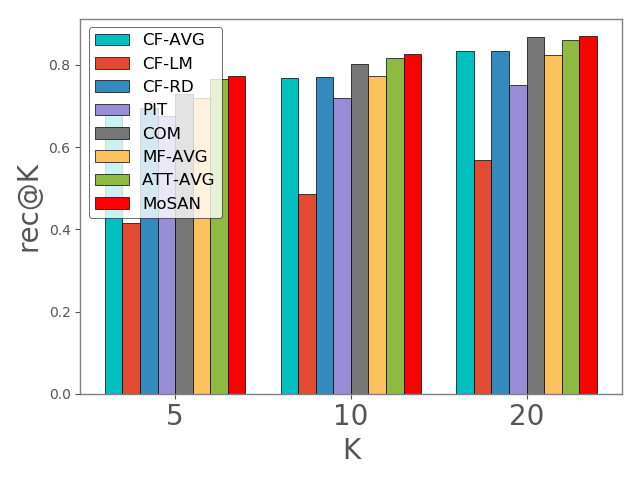}
\centering{Meetup}
\end{minipage}
\begin{minipage}[t]{4.39cm}
\includegraphics[width=4.39cm]{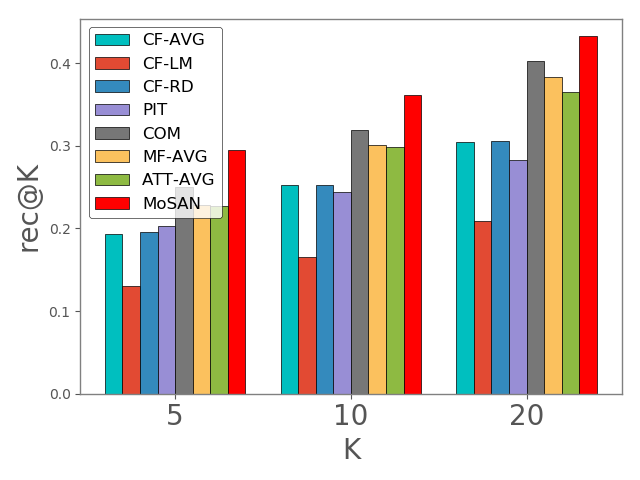}
\centering{Plancast}
\end{minipage}
\begin{minipage}[t]{4.39cm}
\includegraphics[width=4.39cm]{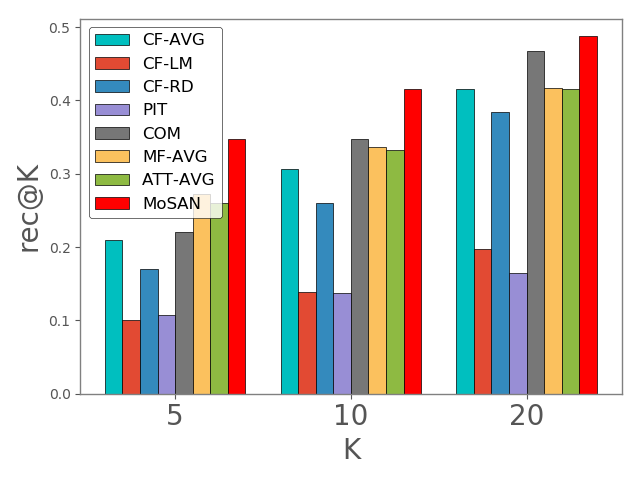}
\centering{MovieLens-Simi}
\end{minipage}
\begin{minipage}[t]{4.39cm}
\includegraphics[width=4.39cm]{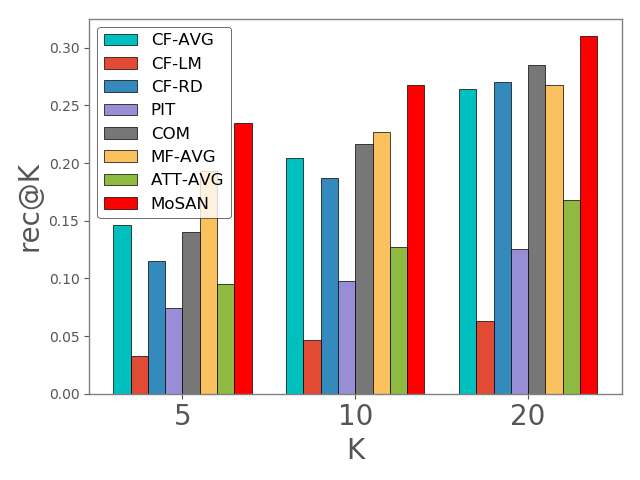}
\centering{MovieLens-Rand}
\end{minipage}
\caption{Performance of Group Recommendation Methods in terms of rec@K $(p<0.0001)$ (\textmd{\textit{Best viewed in color}}).}
\label{fig:rec}
\end{center}
\vspace{-2ex}
\end{figure*}

\begin{figure*}[t]
\begin{center}
\begin{minipage}[t]{4.39cm}
\includegraphics[width=4.39cm]{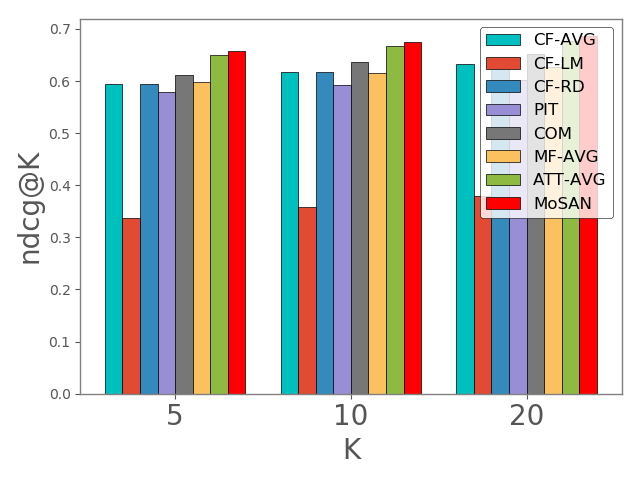}
\centering{Meetup}
\end{minipage}
\begin{minipage}[t]{4.39cm}
\includegraphics[width=4.39cm]{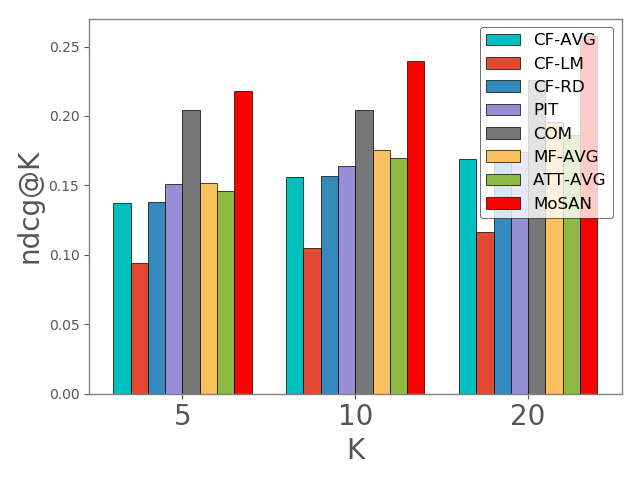}
\centering{Plancast}
\end{minipage}
\begin{minipage}[t]{4.39cm}
\includegraphics[width=4.39cm]{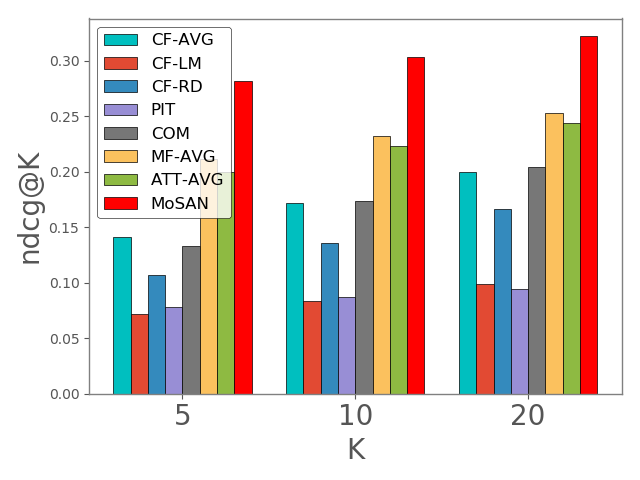}
\centering{MovieLens-Simi}
\end{minipage}
\begin{minipage}[t]{4.39cm}
\includegraphics[width=4.39cm]{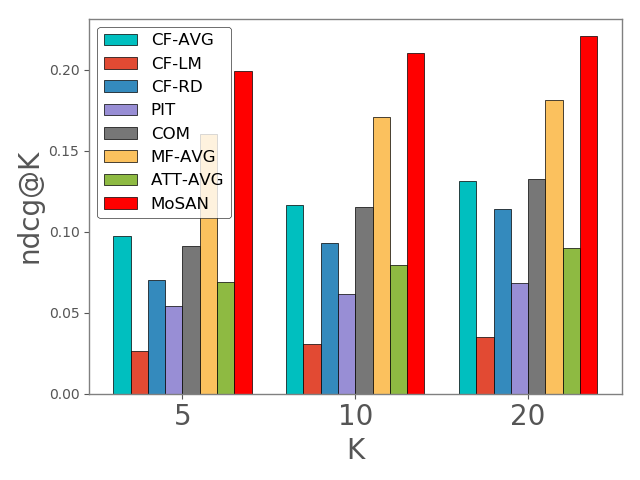}
\centering{MovieLens-Rand}
\end{minipage}
\caption{Performance of Group Recommendation Methods in terms of ndcg@K $(p<0.0001)$ (\textmd{\textit{Best viewed in color}}).}
\label{fig:ndcg}
\end{center}
\vspace{-2ex}
\end{figure*}

Our second dataset is the dataset crawled from the EBSN Meetup,\footnote{https://www.meetup.com/} which is from the work \cite{Pham:2016:GRM:3024719.3024759}. We select the NYC data, which contains events held in New York City, as the dataset for our experiments. Similar to Plancast, we aim to recommend a venue for a given group to host an event. The statistics of this dataset is different from which reported in \cite{DBLP:conf/cikm/LiuTYL12} due to the difference in the period of crawling. 

The final two datasets are obtained from the MovieLens 1M Data.\footnote{https://grouplens.org/datasets/movielens/} The MovieLens 1M Data contains one million movie ratings from over 6,000 users on approximately 4,000 movies. Following the approach in \cite{DBLP:conf/recsys/BaltrunasMR10}, we extract from the MovieLens 1M Data two datasets: MovieLens-Simi and MovieLens-Rand. MovieLens-Simi contains groups with high user-user similarity. We select top 33\% in terms of similarity of all possible pairs to form groups from there. MovieLens-Rand contains groups that are formed without the restrictions above. For a given group in both cases, if every member gives 4 stars or above to a movie, we assume that the movie is adopted by the group. Users in the MovieLens-Simi data are assigned into the same group when they have high inner group similarity, while users in the MovieLens-Rand data are grouped randomly. MovieLens-Simi and MovieLens-Rand groups thereby resemble two typical real life situations: groups can either include people with similar preferences, or form between unrelated people. For example, a group of close friends has high inner group similarity, whereas people on the same bus can be considered a random group.

\begin{table}[t]
	\centering
	\resizebox{\linewidth}{!}{%
		\begin{tabular}{ccccc}
			\hline
			\textbf{Dataset} & \textbf{Plancast} & \textbf{Meetup} & \textbf{\begin{tabular}[c]{@{}c@{}}MovieLens\\ -Simi\end{tabular}} &
			\textbf{\begin{tabular}[c]{@{}c@{}}MovieLens\\ -Rand\end{tabular}} \\ \hline
			\multicolumn{1}{c|}{Total Users} & 41,065 & 42,747 & 5,759 & 5,802 \\
			\multicolumn{1}{c|}{Total Groups} & 25,447 & 13,390 & 29,975 & 54,969 \\
			\multicolumn{1}{c|}{Total Items} & 13,514 & 2,705 & 2,667 & 3,413 \\
			\multicolumn{1}{c|}{Avg. Group Size} & 12.01 & 16.66 & 5.00 & 5.00 \\
			\multicolumn{1}{c|}{\begin{tabular}[c]{@{}c@{}}Avg. Record/User\end{tabular}} & 7.44 & 5.22 & 26.03 & 47.37 \\
			\multicolumn{1}{c|}{\begin{tabular}[c]{@{}c@{}}Avg. Record/Item\end{tabular}} & 1.88 & 4.95 & 11.24 & 16.11 \\
      \hline
		\end{tabular}}
	\caption{Dataset Statistics}
	\label{data_summary}
	\vspace{-4ex}
\end{table}

Table \ref{data_summary} reports descriptive statistics of the four datasets. We randomly split each dataset into training, tuning and testing data with the ratio of $70\%$, $10\%$ and $20\%$ respectively. Note that the groups in our setting are ad-hoc, i.e., it is possible that a group appears only in test data, but not in training data. 

\smallskip\noindent \textbf{Evaluation Metrics.} Following previous work \cite{DBLP:conf/recsys/BaltrunasMR10,DBLP:conf/cikm/LiuTYL12,Pham:2016:GRM:3024719.3024759,DBLP:conf/kdd/YuanCL14}, we evaluate model performance using three widely used evaluation metrics: \textit{precision} (\textit{prec@K}), \textit{recall} (\textit{rec@K}), and normalized discounted cumulative gain (NDCG) (\textit{ndcg@K}). Here $K$ is the number of recommendations. We evaluate recommendation accuracy with $K = \{5, 10, 20\}$. \textit{precision@K} is the fraction of top-$K$ recommendations selected by the group, while \textit{recall@K} is the fraction of relevant items (true items) that have been retrieved in the top $K$ relevant items. We average the \textit{precision@K} and \textit{recall@K} values across all testing groups to calculate \textit{prec@K} and \textit{rec@K}. We also use the \textit{NDCG} metric to evaluate the rankings of true items in the recommendation list. We average the \textit{NDCG} values across all testing groups to obtain the \textit{ndcg@K} metric. For all of the three metrics, a larger metric value indicates better recommendations.

\smallskip\noindent \textbf{Compared Baselines.} We compare with seven state-of-the-art baselines in our experiments: CF-AVG, CF-LM, CF-RD \cite{DBLP:journals/pvldb/Amer-YahiaRCDY09}, PIT \cite{DBLP:conf/cikm/LiuTYL12}, COM \cite{DBLP:conf/kdd/YuanCL14}, MF-AVG and ATT-AVG. 

\begin{itemize}
\item \textbf{User-based CF (CF-AVG, CF-LM, CF-RD)} \cite{DBLP:journals/pvldb/Amer-YahiaRCDY09}: The baselines are standard user-based collaborative filtering methods by integrating pre-defined aggregation strategies which are  averaging strategy (CF-AVG), least-misery strategy (CF-LM) and relevance and disagreement strategy (CF-RD). 




\item \textbf{Personal impact topic model (PIT)} \cite{DBLP:conf/cikm/LiuTYL12}: PIT is an author-topic model. Assuming that each user has an impact weight that represents the influence of the user to the final decision of the group, PIT chooses a user with a relatively large impact score as the group's representative. The selected user then chooses a topic based on her preference, and then the topic generates a recommended item for the group.

\item \textbf{Consensus model (COM)} \cite{DBLP:conf/kdd/YuanCL14}: COM relies on two assumptions: (i) the personal impacts are topic-dependent, and (ii) both the group's topic preferences and individuals' preferences influence the final group decision. 

\item \textbf{Average Matrix Factorization (MF-AVG)}: This baseline is a simplified version of MoSAN and considers the average embedding of all users in the group. All users are weighted equally. We represent a group as $\textit{g} = \sum_i w_i u_i$ where $w_i = \frac{1}{n}$, and we optimize the BPR objective to predict group recommendation scores. 

\item \textbf{Attentive Aggregation (ATT-AVG)}: This baseline is also a simplified version of MoSAN, and represents a group embedding using a vanilla attentive aggregation over all the user embeddings in the group. The user embeddings are optimized with the BPR objective function.

\end{itemize}
\vspace{-0.6ex}
\smallskip\noindent \textbf{Hyperparameter Settings.} For PIT and COM, we tuned the number of topics and kept other hyperparameters as default. With regard to the MF-AVG/ATT-AVG and MoSAN models, we first randomly initialize the parameters using the Gaussian distribution with mean of 0 and standard deviation of 0.05, and then use Adaptive Moment Estimation (Adam) to optimize our objective functions. We also tested the batch size of [128, 256, 512], the learning rate of [0.001, 0.005, 0.01, 0.05, 0.1], and different regularizers of [0.001, 0.01, 0.1, 0]. We empirically set the embedding size of MF-AVG/ATT-AVG and MoSAN with the dimension of 50. We obtain the optimal setting with the batch size of 256, learning rate of 0.001, and regularizers of 0.01. We randomly set dropout rate $\rho=0.5$ and sample 3 negative items for each training instance.  

\begin{table*}[t]
	\centering
		\begin{tabular}{|c|c|c|c|c|c|c|c|c|}
			\hline
			\multirow{2}{*}{\textbf{Model}} & \multicolumn{2}{c|}{\textbf{Meetup}} & \multicolumn{2}{c|}{\textbf{Plancast}} & \multicolumn{2}{c|}{\textbf{\begin{tabular}[c]{@{}c@{}}MovieLens\\ -Simi\end{tabular}}} & \multicolumn{2}{c|}{\textbf{\begin{tabular}[c]{@{}c@{}}MovieLens\\ -Rand\end{tabular}}} \\ \cline{2-9}
			& \textit{rec@5} & \textit{ndcg@5} & \textit{rec@5} & \textit{ndcg@5} & \textit{rec@5} & \textit{ndcg@5} & \textit{rec@5} & \textit{ndcg@5} \\ \hline
			MF-AVG & 0.851962 & 0.632753 & 0.467176 & 0.211513 & 0.321079 & 0.191142 & 0.530135 & 0.274007 \\ \hline
            ATT-AVG & 0.883432 & 0.681998 & 0.439959 & 0.200346 & 0.221269 & 0.099581 & 0.520669 & 0.263710 \\ \hline
			MoSAN & \textbf{0.888513} & \textbf{0.689149} & \textbf{0.502620} & \textbf{0.270515} & \textbf{0.360627} & \textbf{0.230099} & \textbf{0.578485} & \textbf{0.338959} \\ \hline
			$p$-value & $< 0.0001$ & $< 0.0001$ & $< 0.0001$ & $< 0.0001$ & $< 0.0001$ & $< 0.0001$ & $< 0.0001$ & $< 0.0001$ \\ \hline
		\end{tabular} 
        \caption{Performance comparison between MF-AVG and ATT-AVG (Ablation study) on four datasets. Results show that our proposed attention mechanism is significantly better than a standard attention-based aggregation. }
	\label{effect_attention}
	\vspace{-4ex}
\end{table*}

\vspace{-2ex}
\subsection{Overall Performance Comparison (RQ 1)}
\label{sssec:experimental_results}

This subsection compares the recommendation results from MoSAN to those from the baseline models. Figure \ref{fig:prec}, Figure \ref{fig:rec} and Figure \ref{fig:ndcg} report the \textit{prec@K}, \textit{rec@K} and \textit{ndcg@K} values for the four datasets with $K = \{5, 10, 20\}$. We observe from the three figures that:

\begin{itemize}

	\item MoSAN consistently achieves the best performance across all methods, including score-aggregation approaches (CF-AVG, CF-LM, CF-RD) and probabilistic model approaches (PIT, COM).
	\item Although the group information of the two MovieLens datasets is generated manually instead of already observed as Meetup and Plancast, our model can still be able to show the recommendation flexibility in adopting to randomness datasets.
	\item MoSAN and MF-AVG models produce good results in comparison to the previous state-of-the-art probabilistic models. MF-AVG performs better than PIT and COM on the MovieLens-Simi and MovieLens-Rand datasets in terms of \textit{ndcg@K}, but not on the Meetup and Plancast datasets. One explanation is that the simplistic setup of MF-AVG cannot model the complexity of real life group interactions.

\end{itemize}

\begin{figure}[t]
\begin{center}
\begin{minipage}[t]{5.8cm}
\includegraphics[width=5.8cm]{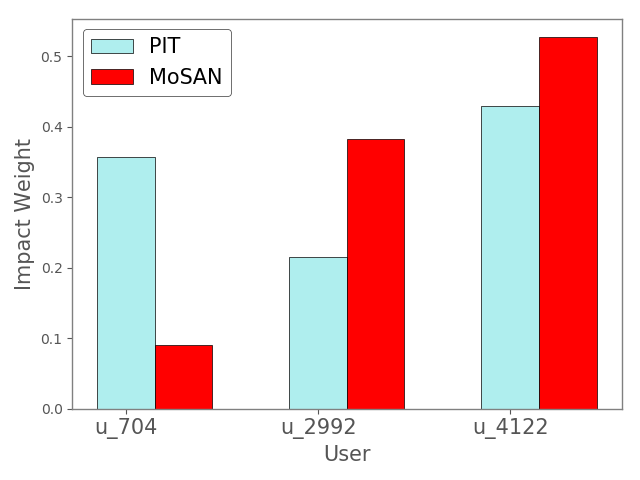}
\centering{(a) Group A}
\end{minipage}
\begin{minipage}[t]{5.8cm}
\includegraphics[width=5.8cm]{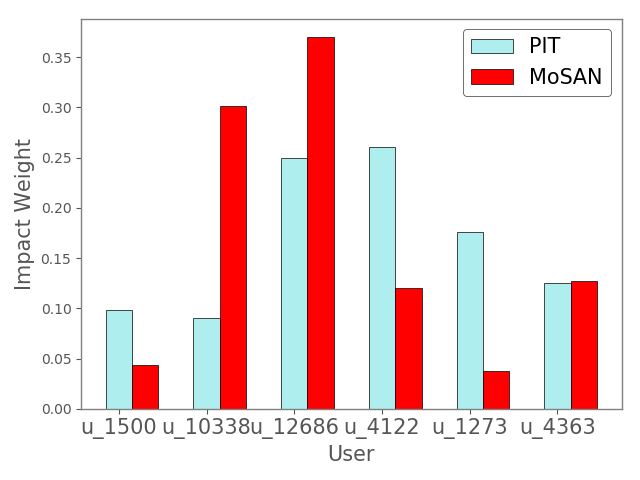}
\centering{(b) Group B}
\end{minipage}
\caption{Attention Weights Learned by PIT and MoSAN.}
\label{fig:visualization}
\end{center}
\vspace{-4ex}
\end{figure}

We observe that there is no obvious winner among the baseline solutions. For each dataset, we define the baseline that has the greatest performance among the proposed ones as \textit{the best baseline} method. The \textit{prec@5} metric values show that MoSAN outperforms the best baseline method significantly by $0.94\%$, $17.91\%$, $27.71\%$, $21.36\%$ on Meetup, Plancast, MovieLens-Simi and MovieLens-Rand, respectively. We observe the same improvements for \textit{rec@5}. In general, MoSAN's recommendations are consistently better than the baseline methods', with the $p$-value less than 0.0001 for all the results and thus statistically significant.

It is noticeable that PIT's performance is not comparable with those of other baseline models. One possible reason is that as a Meetup or Plancast group usually has a large number of participants, many of whom only join a few groups and thus have very limited historical data. Hence, the user impacts learned by PIT for such participants are not reliable. Another possible reason for PIT's poor performance is that the assumptions underlying PIT do not hold in the context of MovieLens data: since MovieLens users select movies independently from one another, there is no representative user in a MovieLens group.

We also observe that MoSAN does not outperform the baselines models on the Meetup dataset as significantly as it does on the other three datasets. One explanation is that since a Meetup group often has few venue options, and group members tend to choose the place they are most familiar with, making it relatively easy to recommend the venue to the group \cite{Pham:2016:GRM:3024719.3024759}. While Meetup users form a big group before hosting an event and choosing the venue, Plancast allow users to follow other users' event calendars and choose to participate in existing events. Plancast groups therefore tend to be more diverse than Meetup groups, and Plancast event venues are not as easily predicted as Meetup event venues.

In addition, one may concern the ability of MoSAN to deal with large groups. However, it should not be an issue as we observe that typically most of groups have fewer than 10 users. For larger groups, we can always reduce the number of users to a smaller number (e.g., 20 users). Then, we can use our model to firstly compute the attention weights of users and remove the users with very small weights. We can always perform this step because in reality, the number of group leaders/experts who participates in making the final decision for the group is always very small. 

In general, MoSAN achieves remarkable recommendation results on the variety of datasets consistently. Our experiments show the flexibility of MoSAN in making group recommendation given different data types.

\subsection{The Role of Attention Mechanism (RQ 2)}
\label{sssec:role_of_attention}

\smallskip\noindent \textbf{Ablation Study.} We further conduct paired $t$-tests on the performances of MoSAN against MF-AVG and ATT-AVG models to verify that MoSAN's improvements over MF-AVG and ATT-AVG are statistically significant at the five percent significance level. While MoSAN employs a user-user attention mechanism to calculate the weights, MF-AVG assigns a normalized constant weight to each group member. Similarly, ATT-AVG is a simple attentive pooling over all users. Even though it also adopts an attention mechanism, it does not consider user-user interactions. We conduct paired $t$-tests on the performance metrics of top-$K$ recommended lists ($K \in [10, 100]$) for MoSAN, MF-AVG and ATT-AVG to see whether MoSAN statistically significantly outperforms MF-AVG and ATT-AVG. We also observe that ATT-AVG does not always outperform MF-AVG, signifying that the standard attention mechanism is insufficient. 

Table \ref{effect_attention} compares the performances of MoSAN and MF-AVG. We observe that the mean pooling strategy of MF-AVG always performs worse than the attention mechanism of MoSAN. The good performance of MF-AVG on MovieLens-Rand data is expected because MovieLens-Rand groups satisfy MF-AVG assumptions: randomly-grouped users in MovieLens-Rand data tend to equally contribute to the groups' final decisions. The reported $p$-values are nominal for all tests, indicating that MoSAN's performances are statistically much stronger than MF-AVG's performances.

\begin{table}[t]
	\centering
	\resizebox{\linewidth}{!}{%
		\begin{tabular}{|c|c|c|c|c|}
			\hline
			\multirow{2}{*}{\textbf{Number of users $K$ to be removed}} & \multicolumn{2}{c|}{\textbf{Meetup}} & \multicolumn{2}{c|}{\textbf{Plancast}} \\ \cline{2-5}
			& MoSAN & PIT & MoSAN & PIT \\ \hline
			$K=1$ & 0.643932 & 0.56832 & 0.197189 & 0.14193 \\ \hline
            $K=3$ & 0.593263 & 0.56690 & 0.118760 & 0.14085 \\ \hline
			$K=5$ & 0.500849 & 0.56424 & 0.081018 & 0.14013 \\ \hline
			$K=7$ & 0.377586 & 0.56238 & 0.058164 & 0.13954 \\ \hline
			$K=9$ & 0.284848 & 0.56163 & 0.046473 & 0.13950 \\ \hline
		\end{tabular} }
        \caption{Performance comparison between MoSAN and PIT on Meetup and Plancast datasets in terms of ndcg@5 by removing top-$K$ high weight users.}
	\label{remove_high_weight_users}
	\vspace{-6ex}
\end{table}

\smallskip\noindent \textbf{Attention Weight Visualization.} As noted previously, an advantage of MoSAN is that the method allows us to calculate the attention weight values for explanation of group recommendation results. Figure \ref{fig:visualization} visualizes the attention weights learned by MoSAN and PIT for two randomly-chosen groups from our experiments. We compare with the weights learned by PIT because similar to MoSAN, PIT is able to learn the personal impact weight for each user \cite{DBLP:conf/cikm/LiuTYL12}. Since COM does not learn the personal impact weight for each user \cite{DBLP:conf/kdd/YuanCL14}, we do not compare MoSAN with COM here.

Figure \ref{fig:visualization}(a) and Figure \ref{fig:visualization}(b) show user attention weights learned by PIT and MoSAN of two randomly-chosen groups that share a user no. 4122 (``$u\_4122$''). Figure \ref{fig:visualization}(a) reports the personal impact weights of three users in the first group (group A). According to both models, the user no. 4122 has the highest weights in group A and therefore dominates group A's decision making. Figure \ref{fig:visualization}(b), however, shows that PIT continues to assume that user no. 4122 is the most influential user in group B, whereas MoSAN is able to detect other users who play more important roles in group B's decision making than user no. 4122 does. While PIT's personal impact parameter cannot differentiate the roles of one user in different groups and thus may fail to recognize influential members in a group, the attention mechanism of MoSAN can capture the dynamic user impacts across groups in group decision making.

\smallskip\noindent \textbf{The importance of high weight users.} In addition, we also conduct another ablation experiment to analyze the importance of users by comparing the recommendation results of MoSAN and PIT with and without high weight users on Meetup and Plancast datasets. We rank users by their impact weight in decreasing order and remove top-$K$ users from group recommendation, and report the change in MoSAN's and PIT's performance, in which $K \in \{1, 3, 5, 7, 9\}$.

Table \ref{remove_high_weight_users} shows the performance comparison between MoSAN and PIT in terms of \textit{ndcg@5} by removing top-$K$ high weight users. It is noticeable that the performance of PIT is stable as the number of removed users increases, whereas the performance of MoSAN drops dramatically. Specifically, the average performance drop of MoSAN for removing every additional two users is $18.15\%$ for Meetup dataset and $29.97\%$ for Plancast dataset, while the performance of PIT is nearly unchanged. The decreasing performance of MoSAN shows that the removed users are important users in the group while the top-ranked users in PIT are not really important users; and thus, the attention mechanism in MoSAN can really capture the dynamic user impacts better than PIT. It reflects the effectiveness of MoSAN in analyzing and weighting high/low impact users in different groups, which is critical for group recommendation systems. All in all, the ablation experiment of the importance of users shows the preciseness of MoSAN over PIT in learning the attentive weight for each user. 

\begin{figure}[t]
\begin{center}
\begin{minipage}[t]{5.8cm}
\includegraphics[width=5.8cm]{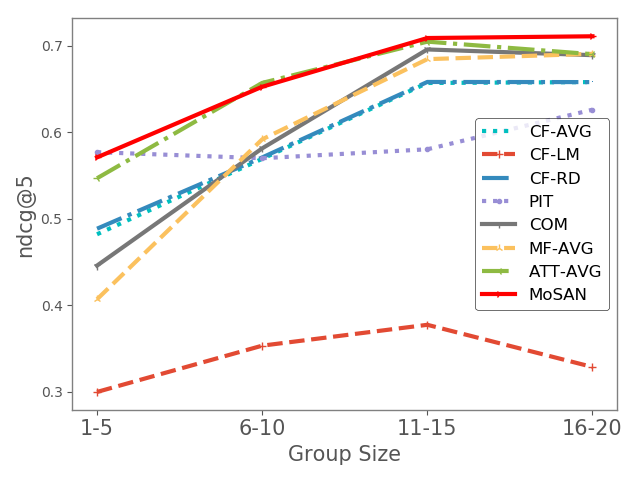}
\centering{(a) Meetup}
\end{minipage}
\begin{minipage}[t]{5.8cm}
\includegraphics[width=5.8cm]{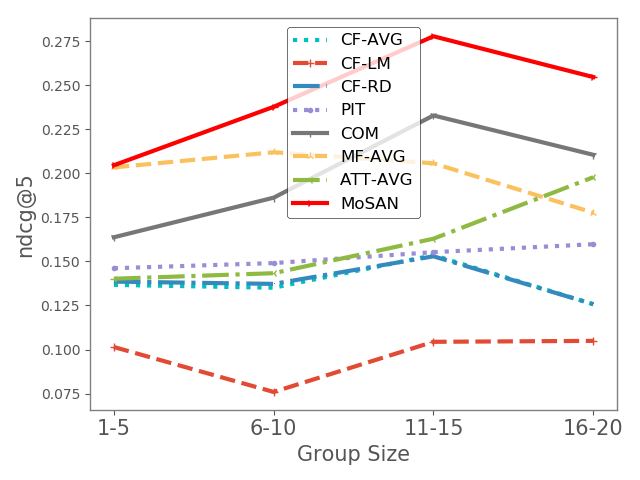}
\centering{(b) Plancast}
\end{minipage}
\caption{Performance on Different Group Sizes (\textmd{\textit{Best viewed in color}}).}
\label{fig:group_size}
\end{center}
\vspace{-5ex}
\end{figure}

\subsection{Model Performances for Different Group Sizes (RQ 3)}
\label{sssec:group_size}

To study the performance of each recommendation method on different group sizes, we run the experiments for four levels of group size (1-5 members, 6-10 members, 11-15 members, and 16-20 members) using Meetup and Plancast groups. We keep the same setting as illustrated for all the models, and classify the groups into bins based on group size. Since the number of groups with more than 20 members is very small, we exclude these groups in this experiment. Figure \ref{fig:group_size} plots the resulting \textit{rec@5} and \textit{ndcg@5} curves. Note that since the group size of the MovieLens-Simi and MovieLens-Rand dataset is fixed, we do not study the different levels of group sizes on these two datasets.

Figure \ref{fig:group_size} shows that MoSAN achieves better performance than other baseline methods across different group sizes. We have the following observations: 1) MoSAN shows clear improvements in recommendations for groups of larger sizes. MoSAN improves $0.59\%$ and $2.93\%$ on the Meetup dataset over the best baseline method for groups of 11-15 members and 16-20 members respectively, while these numbers are $19.30\%$ and $20.97\%$ for the Plancast dataset. This indicates the significance of MoSAN in addressing groups of larger sizes. 2) CF approaches often deliver good performance when the group size is small, as low diversity within a group facilitates smooth aggregations. As the number of members in the group increases, we need relatively complex methods such as probabilistic models or neural networks models to make adequate recommendations.

\section{Conclusion}
\label{sec:conclusion}

In this paper, we propose a new effective neural recommender for group recommendation. The major contributions of MoSAN are that the model not only dynamically learns different impact weights of each given user for different groups, but it also considers the interactions between users in the group. Thus, MoSAN is able to better model the group decision making process. In addition, MoSAN can tell the relative importance of users in a group, and thus make explainable recommendation possible. We conduct extensive experiments on four real-world datasets, demonstrating that MoSAN is capable of outperforming a myraid of strong state-of-the-art baselines.

\bibliographystyle{ACM-Reference-Format}
\bibliography{references}

\end{document}